\documentclass[conference]{IEEEtran}
\IEEEoverridecommandlockouts
\usepackage{cite}
\usepackage{amsmath,amssymb,amsfonts}
\usepackage{algorithmic}
\usepackage[ruled,vlined]{algorithm2e}
\usepackage{placeins}
\usepackage{comment}
\usepackage{graphicx}
\usepackage{float}
\usepackage{placeins} 
\usepackage{textcomp}
\usepackage{xcolor}
\def\BibTeX{{\rm B\kern-.05em{\sc i\kern-.025em b}\kern-.08em
    T\kern-.1667em\lower.7ex\hbox{E}\kern-.125emX}}
\begin{document}

\title{ SD-CGAN: Conditional Sinkhorn Divergence GAN for DDoS Anomaly Detection in IoT Networks\\
}


\author{\IEEEauthorblockN{Henry Onyeka$^{1}$, Emmanuel Samson$^{1}$, Liang Hong $^{1}$,Tariqul Islam $^{2}$, Imtiaz Ahmed$^{3}$, Kamrul Hasan$^{1}$,  }
\IEEEauthorblockA{$^{1}$ Tennessee State University, Nashville, TN, USA \\
$^{2}$ University of Maryland Baltimore County, Baltimore, Maryland, USA \\
$^{3}$Howard University, Washington, DC, USA \\
Email: $\lbrace$\textit{honyeka,esamson,lhong}$\rbrace$@tnstate.edu, 
$\lbrace$\textit{mtislam}$\rbrace$@umbc.edu, $\lbrace$\textit{imtiaz.ahmed}$\rbrace$@howard.edu,
$\lbrace$\textit{mhasan1}$\rbrace$@tnstate.edu}}


\maketitle

\begin{abstract}
The increasing complexity of IoT edge networks presents significant challenges for anomaly detection, particularly in identifying sophisticated Denial-of-Service (DoS) attacks and zero-day exploits under highly dynamic and imbalanced traffic conditions. This paper proposes SD-CGAN, a Conditional Generative Adversarial Network framework enhanced with Sinkhorn Divergence, tailored for robust anomaly detection in IoT edge environments. The framework incorporates CTGAN-based synthetic data augmentation to address class imbalance and leverages Sinkhorn Divergence as a geometry-aware loss function to improve training stability and reduce mode collapse. The model is evaluated on exploitative attack subsets from the CICDDoS2019 dataset and compared against baseline deep learning and GAN-based approaches. Results show that SD-CGAN achieves superior detection accuracy, precision, recall, and F1-score while maintaining computational efficiency suitable for deployment in edge-enabled IoT environments.
\end{abstract}


\section{Introduction}

The evolution of IoT edge networks has enabled ultra-low latency applications such as autonomous vehicles, industrial automation, and mission-critical connected systems. However, this proliferation has significantly increased the attack surface, making IoT infrastructures vulnerable to sophisticated cyber threats like Distributed Denial-of-Service (DDoS) attacks and zero-day exploits. The tight integration of IoT, SDN, and edge computing creates dynamic and heterogeneous traffic patterns that challenge existing anomaly detection systems~\cite{10120907}. Traditional Intrusion Detection System (IDS) solutions, such as rule-based methods and ML models like autoencoders and SVMs, struggle to adapt to these evolving traffic patterns and often fail to detect unseen attacks~\cite{talukder2024machine}. Even deep learning-based IDSs suffer from concept drift, mode collapse, and performance degradation under imbalanced data, which raises the need for more advanced methods \cite{park2022enhanced}. In practical IoT deployments such as smart manufacturing lines, home automation hubs, and sensor-driven industrial monitoring systems, edge devices operate under strict computational budgets. These platforms often lack GPUs, rely on small memory footprints, and require millisecond inference for real-time responses. As a result, lightweight generative intrusion detection models that maintain accuracy under tight resource constraints are essential for feasible on-device deployment.  

Generative Adversarial Networks (GANs) offer a promising direction by modeling high-dimensional network traffic distributions, enabling the detection of both known and zero-day anomalies. GANs consist of a Generator and a Discriminator trained adversarially to synthesize and evaluate data realism~\cite{goodfellow2020generative}. However, conventional GANs often face issues such as instability, mode collapse, and data imbalance when applied to network security~\cite{park2022enhanced}.

To address these limitations, we propose SD-CGAN, a Conditional GAN framework enhanced with Sinkhorn Divergence for stable and geometry-aware training. Conditional GANs incorporate a class-based conditioning mechanism to improve sample specificity and diversity~\cite{mirza2014conditional}. In our model, we leverage Sinkhorn Divergence~\cite{xu2019modeling,chen2022supervised}, an optimal transport-based loss that improves training stability and convergence. Additionally, we employ CTGAN (Conditional Tabular GAN) for synthetic data augmentation to mitigate class imbalance.

The key contributions of this work include:
\begin{itemize}
    \item Design SD-CGAN for anomaly detection in high-dimensional IoT edge traffic, capable of detecting DDoS and zero-day attacks.
    \item Implement CTGAN to generate realistic minority-class samples and address data imbalance in the CICDDoS2019 dataset.
    \item Evaluate SD-CGAN against existing GAN-based anomaly detection models used for network intrusion detection, using the CIC-DDoS2019 dataset.
    \item Optimize SD-CGAN using Sinkhorn Divergence to enhance training stability, reduce mode collapse, and improve computational efficiency for real-time IoT edge network security applications.
\end{itemize}

The rest of this paper is organized as follows: Section II reviews related work; Section III presents the methodology; Section IV discusses the experimental results; and Section V concludes the study.

\section{Related Works}

\subsection{Anomaly Detection in IoT edge Networks}
To enhance anomaly detection in IoT edge networks, various machine learning and deep learning approaches have been proposed \cite{10993583}. Maimo et al.~\cite{maimo2018self} introduced a self-adaptive learning system that adjusts its parameters based on traffic fluctuations, reducing false alarms and improving detection to over 70\%. Hussain et al.~\cite{hussain2020deep} used CNNs to analyze CDRs for DDoS detection in cyberphysical IoT networks, achieving over 90\% accuracy. However, these supervised models depend heavily on labeled data, which is often scarce and quickly outdated. Illiyasu et al.~\cite{iliyasu2022n} trained a GAN solely on benign IoT traffic to define a normal profile, flagging deviations as anomalies and achieving 81.3\% detection. While unsupervised learning improves adaptability, such methods still face high false positive rates if the normal boundary is poorly captured.

\subsection{GAN-Based Anomaly Detection}
GANs have gained traction in IDS research due to their ability to support one-class modeling and synthetic data augmentation. Schlegl et al.~\cite{schlegl2019f} developed AnoGAN for unsupervised image anomaly detection, which inspired similar applications in network domains. Yao et al.~\cite{yao2023scalable} applied BiGAN with Wasserstein distance for IoT-based IDS. Novaes et al.~\cite{novaes2021adversarial} trained a GAN-based IDS on CICDDoS2019, outperforming traditional DL methods like CNN and LSTM. These models show promise in zero-day detection but are often limited by training instability and poor convergence under adversarial loss settings. It is well known that traditional GANs suffer from training instability issues such as vanishing gradients, oscillatory convergence and mode collapse, especially in high-dimensional space and under distribution imbalance \cite{park2022enhanced,chen2022supervised}. These issues arise from the minimax optimization between the generator and discriminator where reaching equilibrium becomes difficult as distributions diverge. When applied in intrusion detection, the instability issue is amplified due to sparse anomalies and multi-modal benign traffic. As a result, GAN-based IDS models often fail to learn a stable representation of benign flows, motivating the need for more geometry-aware loss functions such as Sinkhorn Divergence \cite{park2022enhanced}.

\subsection{Conditional GANs for Imbalanced Data}
Conditional GANs (cGANs) have been applied to address class imbalance by generating targeted synthetic data. Ullah et al.~\cite{ullah2021framework} implemented one-class, binary, and multi-class cGANs to enhance minority-class detection across seven IoT datasets, achieving 98\% accuracy. Ezeme et al.~\cite{ezeme2020design} proposed AD-CGAN to synthesize anomalies and adapt the decision boundary. Benaddi et al.~\cite{benaddi2022adversarial} introduced a cGAN-based framework to generate adversarial traffic that improves CNN-LSTM IDS performance on zero-day threats by 40\%. 

\subsection{Sinkhorn Divergence and Loss Stabilization}
To address instability in GAN training, researchers have explored Optimal Transport (OT)-based metrics. Arjovsky et al.~\cite{arjovsky2017wasserstein} introduced WGAN, using Earth Mover’s Distance to mitigate mode collapse. Cuturi~\cite{cuturi2013sinkhorn} proposed Sinkhorn Divergence, which adds entropy regularization for improved computational efficiency. Genevay et al.~\cite{genevay2018learning} confirmed its utility in generating stable gradients and capturing distribution mismatches. Xu et al.~\cite{xu2020cot} introduced COT-GAN with mixed Sinkhorn divergence to address batch-level convergence. Tien et al.~\cite{tien2023revisiting} applied debiased Sinkhorn divergence to improve image anomaly localization. BEGAN~\cite{berthelot2017began,park2022enhanced} further stabilized training by using autoencoder-based discriminators with equilibrium constraints.

\subsection{Research Gap}

While prior work demonstrates the potential of GANs and cGANs in anomaly detection and data augmentation, several critical gaps remain. Existing GAN-based IDS frameworks often suffer from (i) unstable convergence under adversarial loss dynamics, (ii) high computational cost from discriminator-guided training, and (iii) limited generalization under real-world class imbalance. Moreover, Optimal-Transport-based divergences and conditional generation have been explored separately, but their integration in a Sinkhorn-based discriminator, a geometry-aware one-class anomaly detection framework has not been reported. 
Our work addresses these gaps by proposing SD-CGAN: a Sinkhorn-enhanced cGAN trained solely on benign traffic to learn stable, geometry-preserving latent representations for one-class anomaly detection, supported by CTGAN-based data augmentation, and designed for lightweight deployment in IoT networks.

\begin{figure*}
    \centering
    \includegraphics[width=0.9\linewidth]{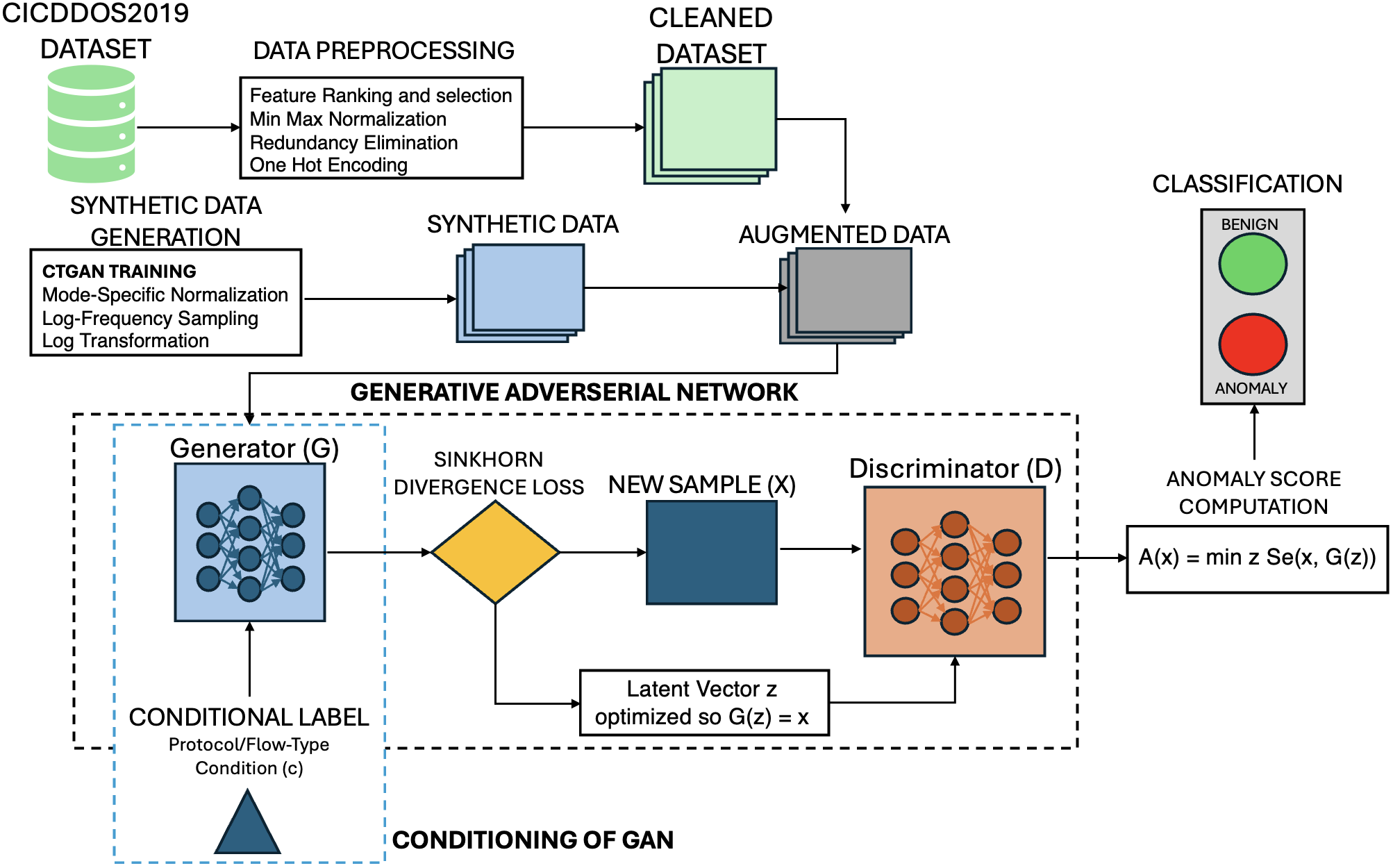}
    \caption{SD-CGAN Model Architecture}
    \label{fig:enter-label}
\end{figure*}

\section{Proposed Methodology}

\subsection{Overview of SD-CGAN Framework}
To address the instability, inefficiency, and poor generalization of traditional GAN-based IDSs, we propose SD-CGAN—a Conditional Generative Adversarial Network trained on benign network flows and optimized using Sinkhorn Divergence. This model is designed to learn the underlying distribution of normal traffic and detect deviations as anomalies, including zero-day attacks. Unlike standard GANs, SD-CGAN relies on a direct statistical comparison between real and generated samples via Sinkhorn Divergence. This stabilizes training, mitigates mode collapse, and allows for lightweight deployment in edge environments.

The SD-CGAN framework involves four major steps: (i) data preprocessing and feature engineering, (ii) synthetic data generation using CTGAN to address class imbalance, (iii) training the conditional generator using Sinkhorn Divergence, and (iv) anomaly scoring based on distributional deviation from benign traffic. Figure 1 illustrates the full system architecture.

\subsection{Conditional GAN Architecture and Training Objectives}
In cGANs, the Generator maps a latent vector \( z \sim \mathcal{N}(0, I) \) and condition \( c \) to a sample \( x \), denoted as:
\[
G : (z, c) \rightarrow x \in \mathbb{R}^n
\tag{1}
\]
where \( G \) learns to produce class-conditioned samples. Here, 
\textit{c} encodes simple flow-type metadata to guide context-aware benign sample generation. Its loss objective:
\[
\mathcal{L}_G = -\mathbb{E}_{z \sim p(z), c \sim p(c)} [\log D(G(z, c) \mid c)]
\tag{2}
\]
encourages realistic synthesis while the Discriminator attempts to classify real vs. generated data:
\begin{multline}
\mathcal{L}_D = -\mathbb{E}_{x \sim p_{\text{data}}, c \sim p(c)} [\log D(x \mid c)] \\
- \mathbb{E}_{z \sim p(z), c \sim p(c)} [\log(1 - D(G(z, c)))]
\tag{3}
\end{multline}
This adversarial game is optimized to approximate the true conditional distribution \( p(x|c) \)~\cite{mirza2014conditional,goodfellow2020generative}.

Conditional GANs are useful for anomaly intrusion detection environments because network traffic is highly imbalanced across classes, with minority attack behaviors exhibiting distinct statistical signatures. By conditioning generation on metadata such as flow type, the model avoids collapsing toward majority patterns and learns class-specific structure. 
\subsection{Sinkhorn Divergence}

Traditional adversarial loss functions such as Jensen-Shannon and KL divergence are often unstable for training GANs, especially in high-dimensional spaces~\cite{pan2020loss}. To address this, we adopt \textit{Sinkhorn Divergence}~\cite{cuturi2013sinkhorn, genevay2018learning}, an entropy-regularized optimal transport (OT) metric that stabilizes optimization and provides meaningful gradient signals even when data distributions have little overlap.

Given two empirical distributions \( \mu \) and \( \nu \) over a metric space, the entropic OT distance is defined as:
\begin{equation}
W_\epsilon(\mu, \nu) = \min_{\pi \in \Pi(\mu, \nu)} \sum_{x,y} \pi(x, y) c(x, y) + \epsilon \mathcal{H}(\pi)
\end{equation}
where \( \Pi(\mu, \nu) \) denotes the set of all couplings (joint distributions) between \( \mu \) and \( \nu \), \( c(x, y) \) is the ground cost function (typically Euclidean), and \( \mathcal{H}(\pi) = -\sum_{x,y} \pi(x,y) \log \pi(x,y) \) is the entropy regularization term.

Since \( W_\epsilon(\mu, \mu) \neq 0 \), we use the debiased Sinkhorn divergence:
\begin{equation}
\mathcal{S}_\epsilon(\mu, \nu) = W_\epsilon(\mu, \nu) - \tfrac{1}{2} W_\epsilon(\mu, \mu) - \tfrac{1}{2} W_\epsilon(\nu, \nu)
\end{equation}

In our SD-CGAN architecture, the Sinkhorn divergence replaces the adversarial loss and the Discriminator. The generator learns by directly minimizing the divergence between mini-batches of real and synthetic data, using the following training loss:
\begin{equation}
\mathcal{L}_{\text{Sink}}(\mu, \nu) = OT_\epsilon(\mu, \nu) - \tfrac{1}{2} \left[ OT_\epsilon(\mu, \mu) + OT_\epsilon(\nu, \nu) \right]
\end{equation}
where \( OT_\epsilon \) is computed efficiently via the \texttt{geomloss} library, allowing scalable mini-batch optimization.

This loss function captures the geometry of the data distribution by measuring the "cost" of transporting mass from generated to real samples using a cost function such as \( c(x, y) = \|f(x) - f(y)\|_2^2 \), where \( f(\cdot) \) maps samples into a learned feature space. As a result, Sinkhorn divergence offers several benefits over classical GAN losses:  
(a) stable and smooth gradient flow during training,  
(b) support for non-overlapping distributions,  
(c) enhanced mode coverage, and  
(d) suitability for unsupervised anomaly detection.

By combining conditional generation with Sinkhorn-based optimization, our model learns a robust representation of benign traffic. During inference, any significant deviation from this distribution, indicated by high Sinkhorn divergence, is flagged as anomalous. This enables SD-CGAN to function effectively as a one-class anomaly detector~\cite{xu2019modeling, chen2022supervised}, while also ensuring training stability and reducing mode collapse.

\subsection{Dataset}
We utilize the CICDDoS2019 dataset~\cite{sharafaldin2019developing}, a flow-based dataset featuring benign and malicious traffic with over 80 statistical and time-based features. Our focus is narrowed to three high-impact, transport-layer exploitative attacks: UDP Flood, UDP-Lag, and SYN Flood, that emulate real-world threats in IoT edge networks due to their bursty, high-volume nature. The dataset's IP-based traffic patterns and protocol-level behavior closely reflect conditions found in edge-deployed IoT systems, making it suitable for evaluating anomaly detection models in this domain.

\subsection{Data Preprocessing}
Preprocessing involved several stages: stratified sampling and schema verification, feature importance ranking via Random Forests~\cite{zhou2020building}, and Pearson correlation analysis for redundancy elimination. High-importance features with low multicollinearity were retained, followed by Min-Max normalization to [0,1] and one-hot encoding of categorical labels. This ensured consistent feature scaling and balance between model interpretability and training efficiency.

\subsection{Synthetic Data Generation with CTGAN}
To mitigate class imbalance, we used CTGAN~\cite{xu2019modeling}, a generative model tailored for mixed-type tabular data. CTGAN applies mode-specific normalization for continuous features and log-frequency sampling for categorical conditioning. It was trained on the selected exploitative classes to generate realistic minority-class samples, which were merged with the original dataset to form a balanced training set. It is also particularly effective for IoT datasets because it models complex dependencies between mixed-type features such as port numbers, flag counts, durations and protocol fields. These attributes often exhibit long-tailed or multimodal behavior, and CTGAN's mode-specific normalization prevents minority modes from being suppressed during training. Therefore, the augmented dataset more accurately reflects real traffic variability.

\subsection{SD-CGAN Training and Anomaly Detection}
SD-CGAN’s generator is a multi-layer perceptron trained solely on benign flows. At each iteration, a latent vector \( z \sim \mathcal{N}(0, I) \) is mapped to a synthetic flow vector \( G(z) \). The Sinkhorn loss is computed between batches of real and generated samples to update the generator. The generator is a 3-layer MLP with hidden dimensions [128, 256, 256] using ReLU activations and a linear output layer. Conditioning is applied by concatenating the latent vector \( z \) with the one-hot encoded flow-type vector \( c \) prior to the first layer.Training uses Adam $2 \times 10^{-4}$, default beta, and batch size 64. During inference, an anomaly score is calculated by minimizing the Sinkhorn divergence between a test sample and its closest generated approximation. Scores above the 95th percentile (based on benign data) are flagged as anomalies. This enables one-class, unsupervised detection of zero-day behaviors without requiring a separate classifier. Algorithm 1 below gives an overview of the model.

SD-CGAN training alternates between generating synthetic flows and minimizing the Sinkhorn Divergence against real benign batches. In each iteration, a latent vector is sampled, concatenated with a conditional label, and passed through the generator to produce a synthetic sample. The Sinkhorn loss computes the transport cost between mini-batches of real and synthetic samples to provide smooth and informative gradients for updating generator parameters. Unlike adversarial GAN training, the OT-based optimization avoids saturation and reduces instability to enable robust representation of benign IoT traffic.
\begin{algorithm}[htbp]
\caption{SD-CGAN Training with Sinkhorn Divergence}
\KwIn{Benign training set $X \in \mathbb{R}^{N \times d}$, latent dimension $d_z$, batch size $B$, learning rate $\eta$, Sinkhorn loss $\mathcal{L}_{\text{Sink}}$}
\KwOut{Trained generator $G_\theta$}
Initialize generator $G_\theta: \mathbb{R}^{d_z} \rightarrow \mathbb{R}^d$ with parameters $\theta$ \\
\For{epoch $= 1$ \KwTo $E$}{
    \For{each mini-batch $X_i \subset X$}{
        Sample latent noise $Z_i \sim \mathcal{N}(0, I) \in \mathbb{R}^{B \times d_z}$ \\
        Generate synthetic batch $\hat{X}_i = G_\theta(Z_i)$ \\
        Compute Sinkhorn loss: \\
        \Indp
        $\mathcal{L}_{\text{Sink}} = \text{OT}_\varepsilon(X_i, \hat{X}_i) - \frac{1}{2} [\text{OT}_\varepsilon(X_i, X_i) + \text{OT}_\varepsilon(\hat{X}_i, \hat{X}_i)]$ \\
        \Indm
        Update $\theta \leftarrow \theta - \eta \nabla_\theta \mathcal{L}_{\text{Sink}}$
    }
}
\Return $G_\theta$
\end{algorithm}

\section{Evaluation and Results}

\subsection{Experimental Setup and Computational Efficiency}
All experiments were conducted on an Intel\textsuperscript{\textregistered} Core\texttrademark~i7-11700KF CPU (16 threads, 32GB RAM, Ubuntu 20.04) without GPU acceleration. The proposed SD-CGAN was trained solely on benign samples from the CICDDoS2019 exploitative attacks subset, with a latent dimension of 100, batch size of 64, Sinkhorn regularization = 0.05, 70/30 data split and learning rate $\eta = 0.0002$ over 10 epochs. The model has a training time of 4.6 seconds, Inference time of 10.9 seconds, and computes anomaly scores under 1 second per sample in batch mode, demonstrating edge-level feasibility. Although no embedded hardware benchmarks are included, the CPU-only training time (4.6s) and sub-second inference show that SD-CGAN is computationally light enough for edge-class devices. This aligns with recent findings showing that edge-class devices such as Raspberry Pi and Jetson Nano can only sustain real-time anomaly detection only when model latency is kept within a tens-of-millisecond range and memory footprints remain under 50MB, as demonstrated in embedded AI evaluations. These studies further CPU-bound execution is typical in IoT deployment where devices lack GPUs and rely lightweight models, thereby supporting SD-CGAN's practical relevance \cite{imteaj2021survey, fi17040179}.  Deep Learning classifier was also evaluated on the same environment, using the same dataset. This model had a training time of approximately 20 seconds and 19 seconds of inference time, further showcasing the efficiency of our SD-CGAN model. The baseline compared metrics are reported from their original publications and may involve different computational settings but are included for reference, evaluation and methodological context. 

Prior GAN and hybrid models such as those by Novaes et al.~\cite{novaes2021adversarial} and Freitas et al.~\cite{de2023unsupervised} required over 157 epochs of adversarial training without reporting total training time, and other hybrid approaches like DFNN-SAE-DCGAN involve deeper stacked modules that incur higher computational overhead. Similarly, the IDS by Freitas et al \cite{de2023unsupervised}, which incorporates TCN and self-attention mechanisms, achieves strong accuracy but relies on GPU-accelerated infrastructure and LSTM-based layers with longer training cycles Furthermore, training only on benign traffic enables robust zero-day detection, as deviations from the learned benign manifold are flagged as anomalies without requiring labeled attack data, hence supporting zero-day attack detection.

\subsection{Evaluation Metrics}
We evaluate SD-CGAN using four standard classification metrics: precision, recall, F1-score, and accuracy \cite{goutte2005probabilistic}. These metrics were computed using confusion matrix values obtained by comparing predicted labels against ground truth annotations.

\subsection{Evaluation of Synthetic Data}
To assess the quality of synthetic samples generated via CTGAN, we performed both qualitative distribution comparison and statistical testing.

1) Distribution Analysis: 
Kernel Density Estimation (KDE) plots for selected features confirmed strong overlap between real and synthetic samples, indicating CTGAN’s ability to capture multi-modal tabular distributions without mode collapse (Figure 2).
\begin{figure}
    \centering
    \includegraphics[width=1\linewidth]{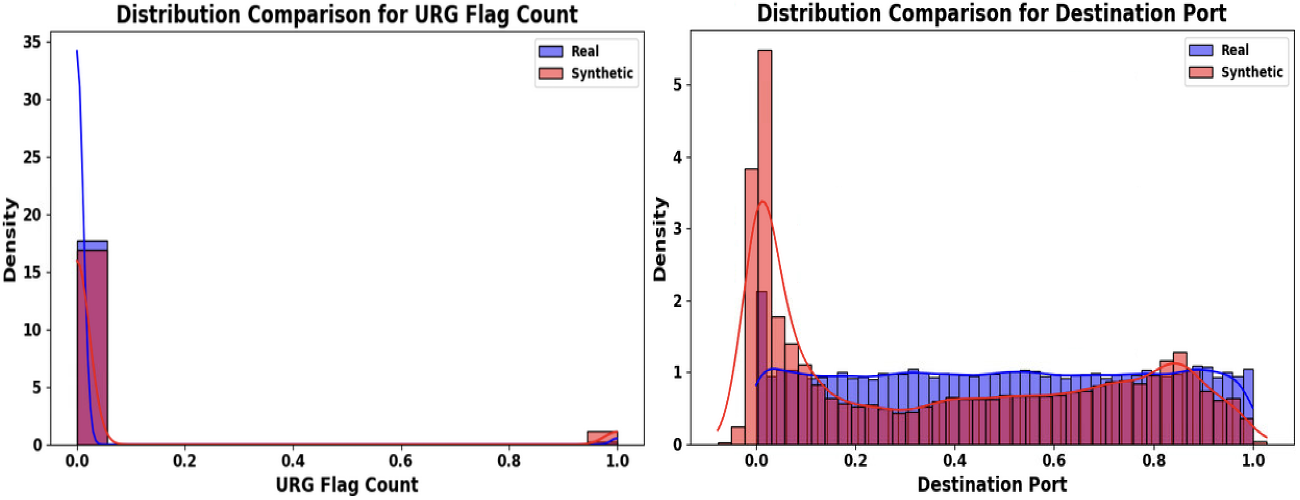}
\end{figure}
\begin{figure}
    \centering
    \includegraphics[width=1\linewidth]{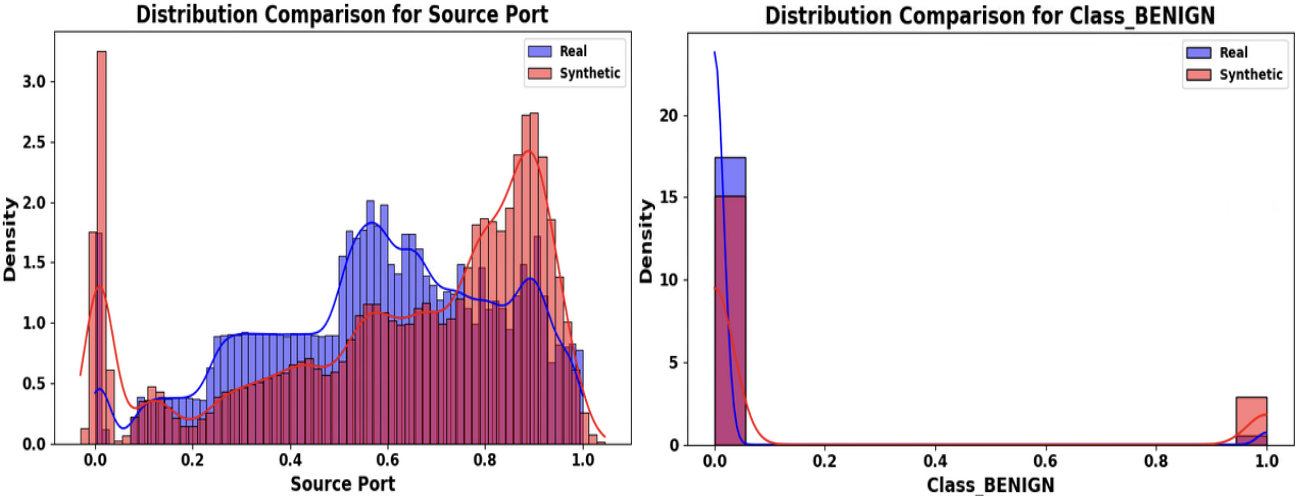}
    \caption{Distribution Comparison Between Real and CTGAN-Generated
Synthetic Samples for Selected Features}
    \label{fig:enter-label}
\end{figure}

2) Kolmogorov–Smirnov Test:
We applied the two-sample KS test to compare real vs. synthetic feature distributions. Table~\ref{tab:ks_top10} shows that several features yielded low KS statistics, indicating high similarity. This supports the use of synthetic data for augmenting minority attack classes without introducing distributional bias \cite{fasano1987multidimensional}. Table 1 presents the KS test results across the top numerical features.

\begin{table}[htbp]
\centering
\caption{Top 10 Features by KS Statistic (Real vs Synthetic)}
\label{tab:ks_top10}
\begin{tabular}{|l|c|c|}
\hline
\textbf{Feature} & \textbf{KS Statistic} & \textbf{p-value} \\
\hline
min\_seg\_size\_forward & 0.0082 & 0.0005 \\
Class\_WebDDoS          & 0.0072 & 0.0032 \\
CWE Flag Count          & 0.0343 & 0.0000 \\
URG Flag Count          & 0.0731 & 0.0000 \\
Source Port             & 0.0985 & 0.0000 \\
Class\_BENIGN           & 0.1764 & 0.0000 \\
Down/Up Ratio           & 0.1402 & 0.0000 \\
Destination Port        & 0.2051 & 0.0000 \\
Inbound                 & 0.2077 & 0.0000 \\
Class\_UDP-lag          & 0.2435 & 0.0000 \\
\hline
\end{tabular}
\end{table}

\subsection{Experimental Results on CICDDoS2019}
SD-CGAN achieved 98.6\% precision, 98.5\% recall, 98.5\% F1-score, and 98.1\% accuracy, outperforming baseline models including CNN, LSTM, RNN-GAN, FID-GAN, and LSTM-FUZZY~\cite{novaes2020long, de2023unsupervised, akana2023optimized}. Figure 3 is a graphical representation of the comparison. Figure~\ref{fig:roc} illustrates the ROC curve with an AUC of 0.9737, and Figure~\ref{fig:confmat} shows the confusion matrix reflecting high classification accuracy.

\begin{table}[htbp]
\centering
\caption{Performance Comparison with Existing Models on CICDDoS2019}
\label{tab:model_comparison}
\begin{tabular}{|l|c|c|c|c|}
\hline
\textbf{Model} & \textbf{Precision} & \textbf{Recall} & \textbf{F1 Score} & \textbf{Accuracy} \\
\hline
LSTM-FUZZY \cite{novaes2020long}             & 97.89\% & 93.13\% & 95.45\% & 95.26\% \\
GAN \cite{novaes2021adversarial}                   & 94.08\% & 97.89\% & 95.94\%  & 94.38\% \\
CNN \cite{novaes2021adversarial}                    & 96.32\% & 86.29\% & 91.02\%  & 94.08\% \\
LSTM \cite{novaes2021adversarial}                  & 90.12\% & 89.43\% & 89.77\%  & 90.29\% \\
RNN-GAN \cite{akana2023optimized}                & 96.80\% & 97.20\% & 97.50\%  & 97.90\% \\
FID-GAN \cite{de2023unsupervised}                & 92.03\% & 92.03\% & 92.03\% & 92.03\% \\
\textbf{SD-CGAN (Proposed)} & \textbf{98.62\%} & \textbf{99.36\%} & \textbf{98.99\%} & \textbf{98.03\%} \\
\hline
\end{tabular}
\end{table}
\begin{figure}[htbp]
    \centering
    \includegraphics[width=0.9\linewidth]{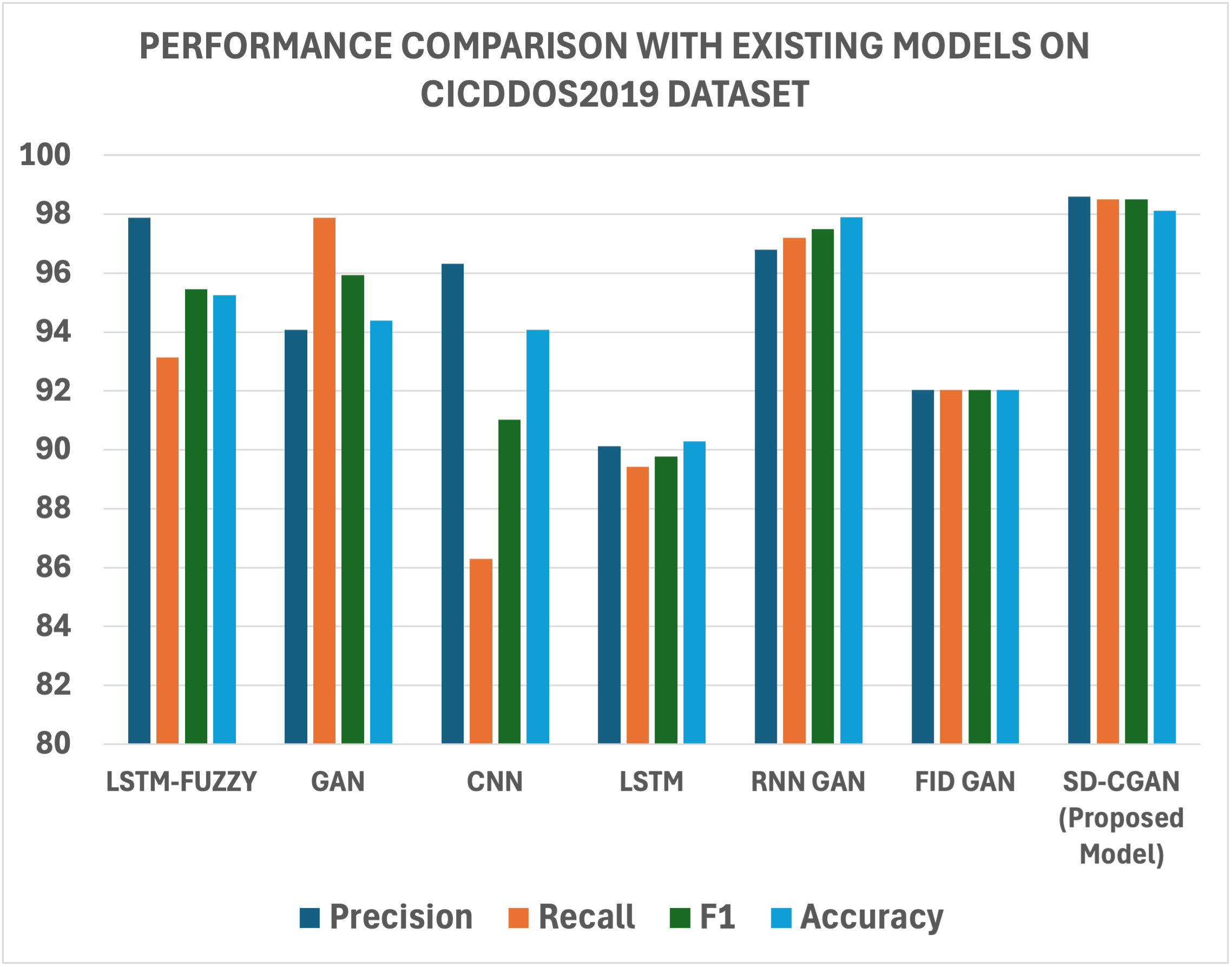}
    \caption{Performance Comparison: SD-CGAN vs Existing Models}
\end{figure}

\begin{figure}[htbp]
    \centering
    \includegraphics[width=0.9\linewidth]{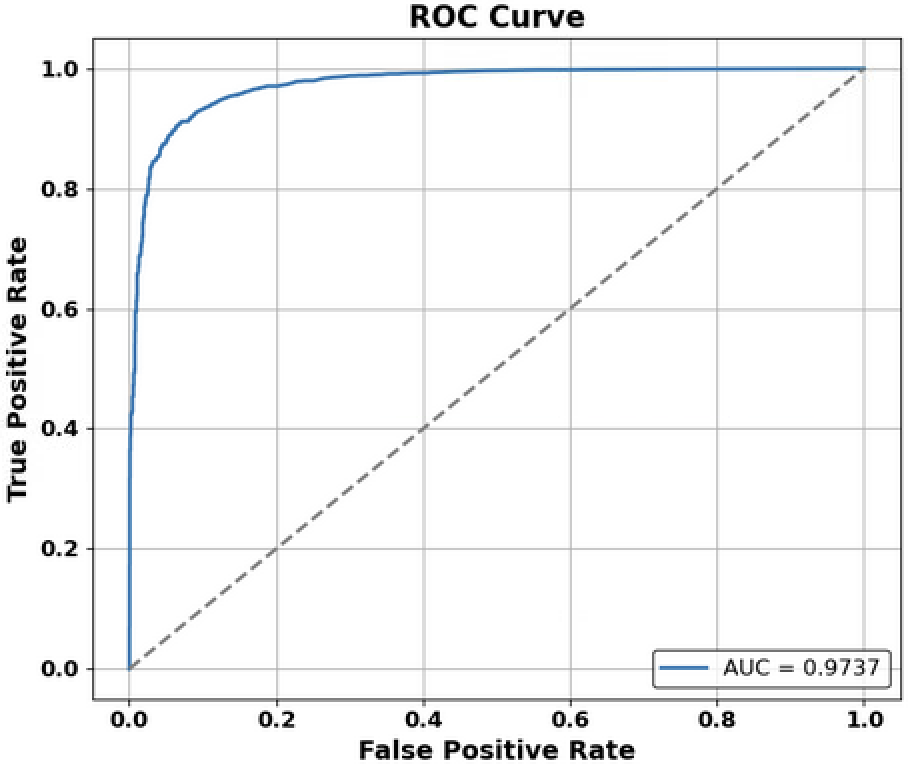}
    \caption{ROC Curve of SD-CGAN (AUC = 0.9737)}
    \label{fig:roc}
\end{figure}

\begin{figure}[htbp]
    \centering
    \includegraphics[width=0.7\linewidth]{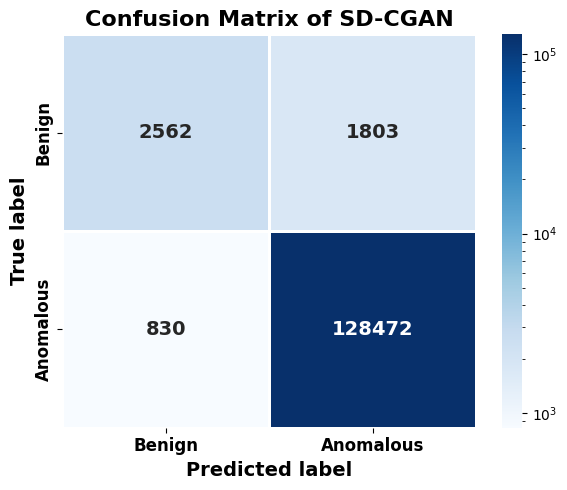}
    \caption{Confusion Matrix of SD-CGAN }
    \label{fig:confmat}
\end{figure}

Also, we validated the original SD-CGAN model for zero-day attack detection by training on benign samples and testing against perturbed attack flows from the existing attack category in the dataset. The model achieved a 98.37\% accuracy with near-perfect performance across all metrics, thereby confirming its ability to flag previously unseen threats as anomalies. An ablation comparison was conducted by training a GAN using the standard Jensen-Shannon divergence under the same environment, and SD-CGAN achieved a higher accuracy (98.62\%) compared to the GAN's (95.04\%), which further confirms the contribution of Sinkhorn Divergence.

\subsection{Training Stability and Convergence}
To evaluate training dynamics, we monitored Sinkhorn loss across epochs. Unlike adversarial losses, the Sinkhorn objective exhibited a smooth decline from 0.59 to 0.21, with no oscillations or collapse. Traditional GANs are known to suffer from mode collapse, and this is modeled in Figure 6. A Deep Learning Classifier which we evaluated can be seen, in Figure 6, suffered from a steep and noisy loss trajectory. This validates geometry-aware divergence as a stabilizer and confirms convergence toward meaningful transport mappings, mitigating mode collapse.

\begin{figure}[htbp]
    \centering
    \includegraphics[width=0.9\linewidth]{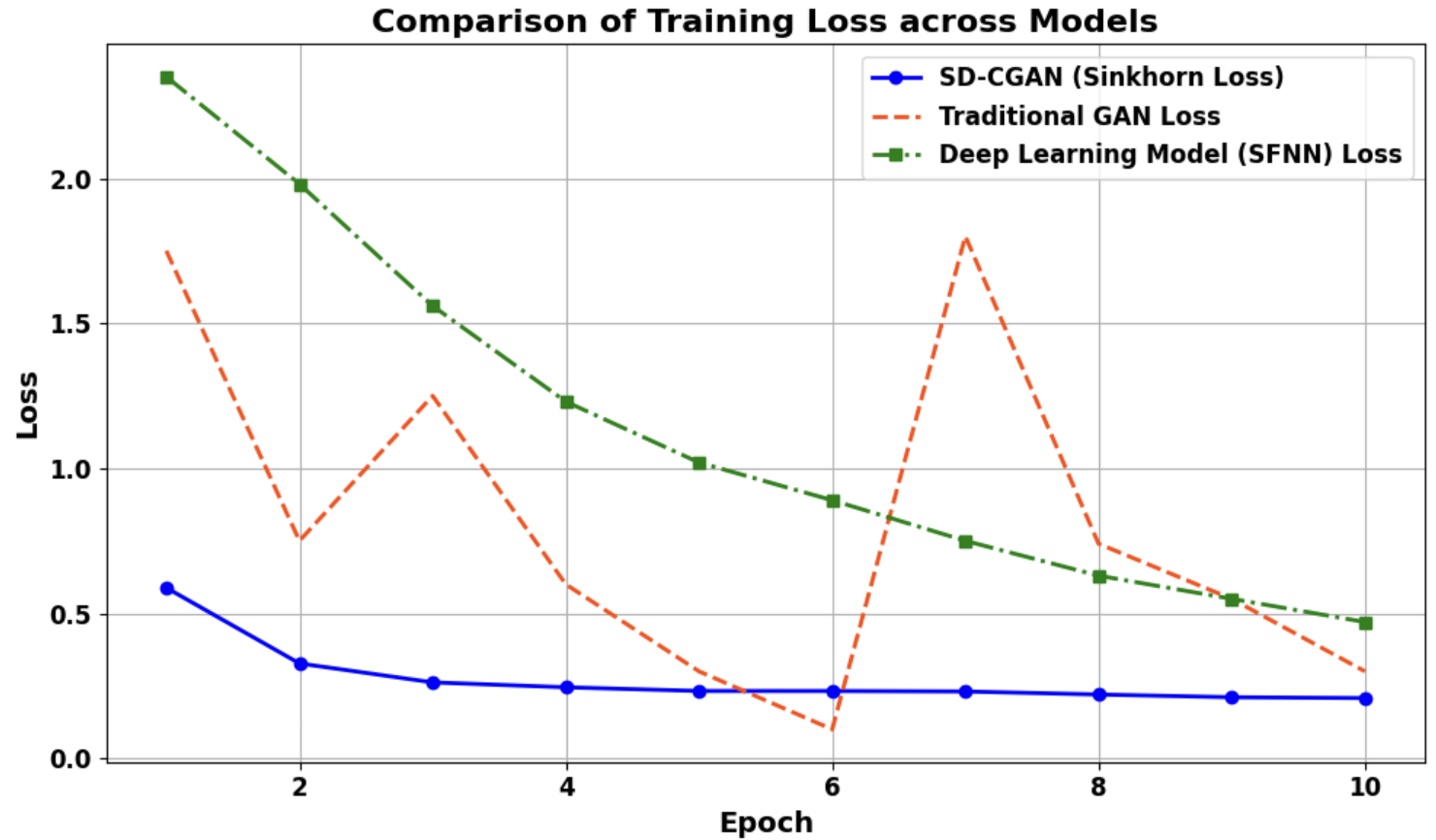}
    \caption{Comparison of Training loss of SD-CGAN against traditional GAN
which suffers mode collapse}
    \label{fig:losscurve}
\end{figure}

\FloatBarrier
\section{Conclusions and Future Works}

This paper introduces SD-CGAN, a cGAN enhanced with Sinkhorn Divergence for anomaly detection in IoT edge networks. Trained exclusively on the CICDDoS2019 dataset with CTGAN-based augmentation, it effectively detects exploitative DDoS attacks under class imbalance. Experiments on the CICDDoS2019 dataset show that SD-CGAN surpasses existing models in accuracy, precision, recall, and F1-score, making it a strong candidate for real-time edge deployment. Mixed-attack evaluation was not included as this study focused on three high-impact transport-layer DDoS attacks to maintain a consistent and controlled experimental scope. Further works include extending SD-CGAN to multi-class scenarios in newer IoT-intensive datasets such as CICIoT2023. Generalizing SD-CGAN across additional IoT traffic datasets will also help evaluate robustness under diverse network scenarios. Explainable AI (XAI) will also be implemented in future works to interpret SD-CGAN's anomaly scoring and false negative behavior, provide feature-level explanations, as certain IoT traffic attributes may exhibit non-linear dependencies that influence optimal-transport-based modelling, and further justify zero-day generalization \cite{10464798}. While SD-CGAN demonstrated strong performance under perturbed zero-day conditions, evaluating generalization to entirely new attack families is an important direction for future work. 

\section*{Acknowledgment}

This work is supported in part by the U.S. Department of Energy (DOE) under Award DE-NA0004189 and the National Science Foundation (NSF) under Award numbers 2409093 \& 2219658.

\bibliographystyle{ieeetr}
\bibliography{references}
\end{document}